\documentclass[a4paper,fleqn]{cas-dc}

\usepackage[authoryear,longnamesfirst]{natbib}
\usepackage{amsmath,amssymb,amsfonts}
\usepackage{booktabs}
\usepackage{multirow}
\usepackage{array}
\usepackage{graphicx}
\usepackage{xspace}

\newcommand{\MoVP}{Mixture of View Projector\xspace}

\def\tsc#1{\csdef{#1}{\textsc{\lowercase{#1}}\xspace}}
\tsc{WGM}
\tsc{QE}
\begin{document}
\let\WriteBookmarks\relax
\def\floatpagepagefraction{1}
\def\textpagefraction{.001}

\shorttitle{SkillMoV for multi-view proficiency estimation}
\shortauthors{E. Bianchi and A. Liotta}

\title[mode=title]{SkillMoV: Mixture-of-View Routing with Prototype-Conditioned Gating for Unified Multi-View Proficiency Estimation}

\author[1]{Edoardo Bianchi}[orcid=0000-0002-0963-9543]
\cormark[1]
\ead{edbianchi@unibz.it}
\credit{Conceptualization, Methodology, Software, Validation, Formal analysis, Investigation, Data curation, Writing -- original draft, Visualization}

\author[1]{Antonio Liotta}[orcid=0000-0002-2773-4421]
\ead{antonio.liotta@unibz.it}
\credit{Conceptualization, Methodology, Supervision, Funding acquisition, Writing -- review and editing}

\affiliation[1]{organization={Free University of Bozen-Bolzano},
            addressline={Faculty of Engineering, Via Bruno Buozzi 1},
            city={Bozen-Bolzano},
            postcode={39100},
            country={Italy}}

\cortext[1]{Corresponding author}

\begin{abstract}
Estimating human proficiency from video is a key challenge for automated skill assessment, with applications in sports coaching, music pedagogy, surgical training, and workplace learning. Existing approaches often focus on individual scenarios or rely on shared multi-view aggregation, limiting their ability to adapt to heterogeneous camera viewpoints and activity domains. We introduce SkillMoV, a unified, parameter-efficient framework for multi-scenario proficiency estimation from synchronized multi-view video. At its core, SkillMoV introduces a \textit{Mixture-of-View Projector} (MoVP), which adapts the mixture-of-experts paradigm to camera-specific view features rather than generic tokens or samples. MoVP is composed of four stages: (i) a \textit{Mixture-of-View} soft router with twelve expert MLPs that learns view-dependent expert preferences without camera-identity supervision; (ii) cross-view attention to align synchronized cameras; (iii) learnable prototype anchoring to condition the representation on class-level reference vectors; and (iv) a prototype-conditioned gated projection that produces the final skill embedding. We evaluate SkillMoV on EgoExo4D across six skill domains and three separately trained view configurations: Ego, Exos, and Ego+Exos. SkillMoV reaches $50.17\%$ overall accuracy in the Exos setting with a single model trained jointly across all scenarios, surpassing the strongest reported Exos result among the compared methods by $3.57$ percentage points. In Ego+Exos, SkillMoV remains close to the best reported result in that setting ($47.63\%$ versus $48.20\%$). Ablations on the selected Exos configuration validate each component: MoV routing contributes $+6.61$ pp over attentive aggregation, cross-view attention $+4.92$ pp, prototype anchoring $+4.07$ pp, and stochastic view dropout $+3.90$ pp. Through LoRA adaptation, SkillMoV trains only $23.32\%$ of its parameters and adds limited measured overhead relative to a LoRA-only baseline.
\end{abstract}


\maketitle

\section{Introduction}
\label{sec:introduction}

Estimating human proficiency from video is a central problem in action quality assessment and skill understanding. Unlike action recognition, where the goal is to identify \emph{what} activity is being performed, proficiency estimation asks \emph{how well} the activity is executed. The relevant evidence often lies in subtle visual cues such as posture, timing, coordination, smoothness, body-object interaction, and task-specific technique. This makes proficiency estimation important for applications such as sports coaching, music pedagogy, rehabilitation, surgical training, and workplace learning, but also substantially more challenging than coarse action classification.

A practical skill-assessment model should generalize across activities and make effective use of multi-view observations. Existing approaches often address proficiency estimation with scenario-specific models or shared aggregation schemes. Scenario-specific training can produce strong results in individual domains, but it limits scalability and requires maintaining separate models or tuning recipes for different activities. Shared multi-view aggregation, on the other hand, can underuse synchronized camera geometry: different cameras may expose different aspects of the same execution, and forcing all views through the same transformation can dilute view-specific proficiency cues.

We introduce SkillMoV, a parameter-efficient framework for unified multi-scenario proficiency estimation from synchronized multi-view video. The key idea is to adapt mixture-of-experts routing to the structure of multi-camera observations. In a standard mixture-of-experts layer, routing is typically applied to generic tokens or samples. In contrast, our \textit{Mixture-of-View} (MoV) routing treats camera-specific view features as the routed units: each synchronized view can select a soft mixture of expert transformations, while the expert pool remains shared across cameras, clips, and activity domains. This turns MoE into a view-conditioned representation mechanism for multi-camera skill assessment, without requiring camera-identity supervision or scenario-specific heads.

SkillMoV builds on a LoRA-adapted TimeSformer backbone and introduces a hierarchical \textit{Mixture-of-View Projector} (MoVP). The projector first applies MoV routing to per-view features, then uses cross-view attention to align synchronized cameras before aggregation. It further conditions the representation with learnable prototype anchors and a prototype-conditioned gated projection. The full model is trained end-to-end with class-balanced cross-entropy, keeping the training objective simple while allowing the projector to learn view-dependent and class-conditioned representations.

We evaluate SkillMoV on the EgoExo4D proficiency estimation benchmark~\citep{grauman2024egoexo4d}, using unified models trained jointly across all skill domains and evaluated under Ego, Exos, and Ego+Exos view configurations. SkillMoV obtains its strongest result in the Exos setting, reaching $50.17\%$ overall accuracy and surpassing the strongest reported Exos result among the compared methods by $3.57$ percentage points. In Ego+Exos, it remains close to the best reported result in that setting ($47.63\%$ versus $48.20\%$). Ablations show that MoV routing is the largest contributor, followed by cross-view attention, prototype anchoring, and stochastic view dropout. Routing analysis further shows that all experts remain active and that soft routing preferences vary across cameras and scenarios without explicit camera or scenario supervision.

This work makes the following contributions:
\begin{itemize}
    \item We propose \textbf{SkillMoV}, a parameter-efficient framework for unified proficiency estimation from synchronized multi-view video.
    \item We introduce \textbf{Mixture-of-View (MoV) routing}, which adapts mixture-of-experts routing to camera-specific view features and learns soft view-dependent expert preferences without camera-identity supervision.
    \item We design the \textbf{\MoVP{}} (MoVP), a hierarchical projector combining MoV routing, cross-view attention, prototype anchoring, and prototype-conditioned gated projection.
    \item We provide a systematic evaluation on EgoExo4D across view configurations, architecture ablations, loss variants, temporal sampling, expert scaling, routing behaviour, and efficiency, showing that SkillMoV improves the Exos comparison while remaining parameter-efficient.
\end{itemize}

\section{Background and Related Work}
\label{sec:related_work}

\subsection{Action quality assessment and proficiency estimation}

Action Quality Assessment (AQA) studies how well an action is executed, complementing action recognition, which focuses on what action is performed~\citep{zhou2024aqasurvey}. Early AQA methods relied on handcrafted or domain-specific descriptors, while deep approaches introduced spatio-temporal representations such as C3D~\citep{tran2015c3d} and I3D~\citep{carreira2017kinetics}. More recent methods use transformer backbones to model long-range temporal dependencies~\citep{bertasius2021timesformer,liu2022videoswin} and task-specific objectives for fine-grained score prediction, ranking, or contrastive regression~\citep{parmar2019mtl,yu2021core,xu2022gdlt,xu2022finediving}. Despite this progress, many AQA systems are designed for a single activity domain, such as diving, skating, or gymnastics, and rely on domain-specific supervision or scoring conventions. This limits their direct applicability to unified proficiency estimation across heterogeneous skills.

EgoExo4D~\citep{grauman2024egoexo4d} broadens this setting by providing synchronized egocentric and exocentric videos of skilled activities, together with benchmark tasks including proficiency estimation. The proficiency-estimation task goes beyond action recognition by aiming to infer the participant's skill level, with synchronized egocentric and exocentric videos enabling different view configurations. This creates a unified multi-domain setting in which a model must handle diverse activities and variable camera configurations. It also motivates models trained across proficiency domains rather than scenario-specific specialists.

\subsection{Unified multi-view proficiency estimation}

Recent EgoExo4D proficiency-estimation methods move from scenario-specific baselines toward unified multi-view models. SkillFormer~\citep{bianchi2025skillformer} introduces a parameter-efficient TimeSformer-based architecture with LoRA adaptation and a CrossViewFusion module that combines egocentric and exocentric features using cross-attention and gating. PATS~\citep{bianchi2025pats} focuses on temporal sampling, preserving continuous movement segments that contain proficiency-relevant execution patterns. ProfVLM~\citep{bianchi2026profvlm} reformulates proficiency estimation as a generative video-language task, predicting skill labels while producing expert-style natural-language feedback through an attentive gated projector and a LoRA-tuned language model.

SkillMoV is positioned within this unified multi-view line, but differs in how it handles synchronized camera features. Prior unified methods primarily fuse views through shared attention, gating, or projection modules. In contrast, SkillMoV introduces a view-conditioned expert-routing mechanism: each camera-specific feature can use a different soft mixture of expert transformations, while the expert pool remains shared across views, clips, and scenarios. This design targets the central challenge of multi-camera proficiency estimation: different viewpoints may expose different skill cues, but the model should still share capacity across heterogeneous activities.

\subsection{Video transformers, temporal sampling, and parameter-efficient adaptation}

Video transformers are widely used for video understanding because they can model spatio-temporal dependencies over frame sequences. TimeSformer~\citep{bertasius2021timesformer} factorizes space-time attention and provides a strong backbone for video classification, while Video Swin~\citep{liu2022videoswin} introduces shifted-window attention for efficient hierarchical video modeling. In proficiency estimation, temporal sampling is also critical: uniform sampling is efficient but may miss or dilute short skill-discriminative moments. PATS~\citep{bianchi2025pats} addresses this issue by preserving temporally coherent movement segments, showing that sampling strategy can strongly affect skill-assessment performance.

Because multi-view video multiplies the cost of backbone inference, parameter-efficient adaptation is important. LoRA~\citep{hu2022lora} freezes the pretrained backbone and learns low-rank adaptation matrices, reducing the number of trainable parameters while retaining adaptation capacity. Recent EgoExo4D proficiency systems use LoRA to adapt TimeSformer-based or video-language models efficiently~\citep{bianchi2025skillformer,bianchi2025pats,bianchi2026profvlm}. SkillMoV follows this parameter-efficient paradigm: the TimeSformer backbone is LoRA-adapted, while the main modeling capacity is concentrated in a lightweight multi-view projector.

\subsection{Mixture-of-experts and view-conditioned fusion}

Mixture-of-Experts (MoE) models route inputs through specialized subnetworks, increasing model capacity while allowing conditional computation~\citep{shazeer2017outrageously,fedus2022switch}. In standard MoE formulations, the routed units are usually tokens, samples, or intermediate representations. SkillMoV adapts this principle to synchronized multi-camera video by treating camera-specific view features as the routing units. We refer to this formulation as \textit{Mixture-of-View} (MoV) routing.

The distinction is important for multi-view proficiency estimation. A shared fusion module must transform all camera features through the same mapping before aggregation, even though different viewpoints may emphasize different cues such as posture, limb extension, hand-object interaction, or body-environment contact. MoV instead lets each view feature select a soft mixture over expert MLPs. The routing remains shared and soft rather than assigning a fixed expert to each camera. This allows view-dependent expert preferences to emerge without camera-identity supervision, while the load-balancing regularizer prevents collapse to a small subset of experts.

\subsection{Ordinal and prototype-conditioned representations}

Proficiency labels are ordinal: adjacent levels such as Early Expert and Intermediate Expert are more related than distant levels such as Novice and Late Expert. Ordinal regression methods such as CORAL~\citep{cao2020coral} enforce rank consistency through cumulative binary classifiers. Prototype-based representation learning provides a complementary view by associating classes with reference vectors in embedding space, as in prototypical networks~\citep{snell2017prototypical}. Contrastive objectives, including supervised contrastive learning~\citep{khosla2020supervised}, further structure embeddings by encouraging semantically related samples to cluster.

SkillMoV incorporates these ideas architecturally rather than relying on an explicit ordinal or contrastive auxiliary loss in its best configuration. It learns one prototype vector per proficiency level and uses cosine similarities to these prototypes to condition the final gated projection. However, the prototypes are optimized end-to-end for classification and are not constrained to form a monotonic ordinal scale. We therefore interpret them as learnable conditioning anchors rather than calibrated ordinal reference points. Auxiliary CORAL, SupCon, and prototype-classification losses are evaluated in ablations, but class-balanced cross-entropy remains the most stable objective for the selected configuration.

\section{Proposed methodology}
\label{sec:method}

\subsection{Problem formulation}
Let $x \in \mathbb{R}^{B \times V \times T \times C \times H \times W}$ denote a batch of synchronized multi-view video clips, where $B$ is the batch size, $V$ is the number of input cameras, $T$ is the number of sampled frames per clip, $C=3$, and $H=W=224$. We evaluate $V=1$ for Ego, $V=4$ for Exos, and $V=5$ for Ego+Exos. The goal is to predict a proficiency label $y \in \{0,1,2,3\}$, corresponding to Novice, Early Expert, Intermediate Expert, and Late Expert respectively. SkillMoV first extracts a per-view feature vector from each clip and subsequently transforms the resulting set of view features into a single skill embedding used for classification.

\subsection{Architecture overview}
\begin{figure*}[t]
    \centering
    \includegraphics[width=0.95\linewidth]{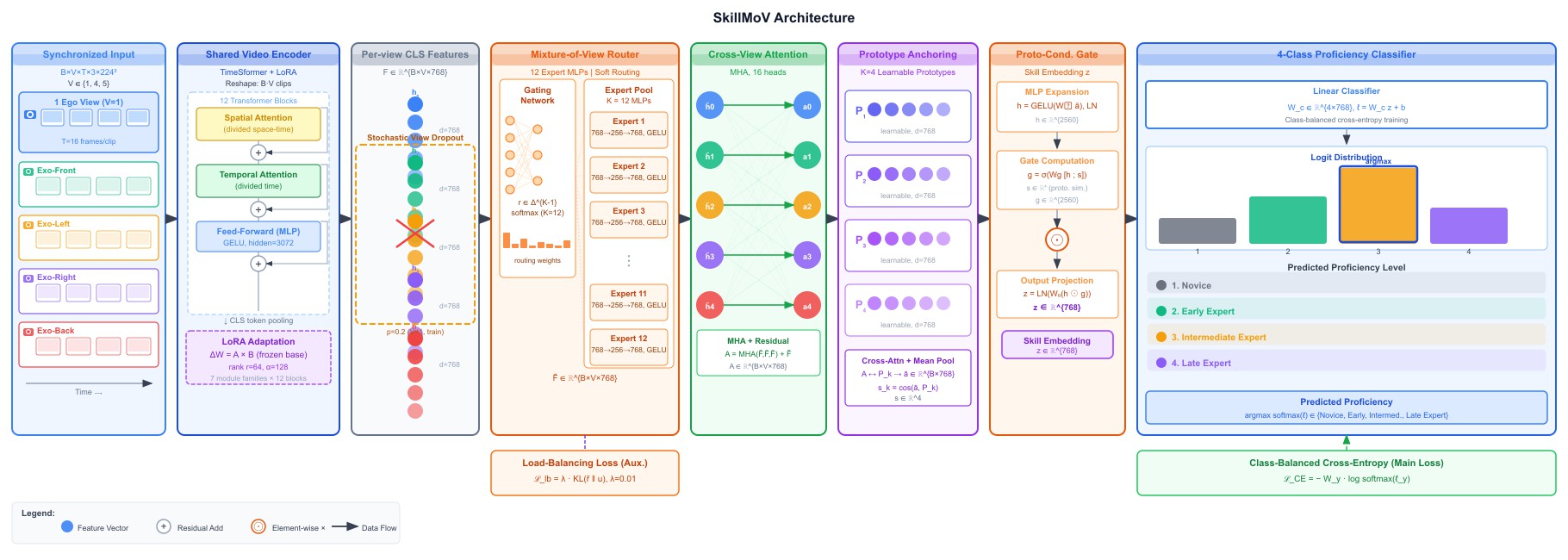}
    \caption{SkillMoV architecture. A shared TimeSformer+LoRA backbone feeds the MoV-Projector (MoVP), which performs Mixture-of-View routing, cross-view attention, prototype anchoring, and prototype-conditioned gated projection before the final proficiency classification head.}
    \label{fig:architecture}
\end{figure*}

Figure~\ref{fig:architecture} summarizes the SkillMoV pipeline. The input tensor is reshaped into $B \cdot V$ clips and passed through a shared TimeSformer-base backbone. The output CLS-pooled features are reshaped into $F \in \mathbb{R}^{B \times V \times 768}$. During training, stochastic view dropout randomly removes a subset of views in the multi-camera configurations while retaining at least two cameras. For the Ego-only configuration ($V=1$), view dropout is disabled and the single egocentric stream is always retained. The remaining features are processed by the \MoVP{} (MoVP), which sequentially performs MoV routing (Section~\ref{sec:mov}), cross-view attention (Section~\ref{sec:cross_view}), prototype anchoring against $C{=}4$ learnable per-class anchors (Section~\ref{sec:proto_anchor}), and prototype-conditioned gated projection (Section~\ref{sec:proto_gate}). A final linear classifier maps the resulting skill embedding to four logits. The whole module is trained end-to-end with class-balanced cross-entropy; we report the loss-function ablation that motivates this choice in Section~\ref{sec:loss_ablation}.

\subsection{Visual Features Extraction}
We employ TimeSformer \citep{bertasius2021timesformer} pre-trained on Kinetics-600 \citep{carreira2017kinetics} as the video backbone. The model contains 12 transformer blocks with hidden dimension 768 and divided space-time attention. Temporal position embeddings are interpolated from 8 to 16 frames at initialization. To adapt the backbone efficiently, we apply LoRA with rank $r=64$, scaling factor $\alpha=128$, and dropout 0.1 to seven module families across all 12 blocks: spatial QKV projections, spatial output projections, temporal QKV projections, temporal output projections, feed-forward intermediate projections, feed-forward output projections, and temporal dense layers. All base backbone weights are frozen; only the LoRA adapters and the projector are trainable.

\subsection{Stochastic view dropout}
Multi-view models can overfit to a dominant camera or degrade when one view is noisy. To regularize cross-view learning in multi-camera settings, we apply stochastic view dropout during training: with probability $p=0.2$, we retain a random subset of views while ensuring that at least two views remain available. This rule is applied to Exos ($V=4$) and Ego+Exos ($V=5$). For Ego-only ($V=1$), view dropout is disabled. This encourages the model to infer proficiency from partial but complementary observations, rather than relying on any single dominant view. The ablation study shows that disabling this mechanism in the selected Exos configuration reduces accuracy from 50.17\% to 46.27\% (Table~\ref{tab:architecture_ablation}).

\subsection{Mixture-of-View routing}
\label{sec:mov}
Mixture-of-View (MoV) adapts the mixture-of-experts principle to synchronized multi-camera video. In a standard MoE layer, routing typically selects experts for generic tokens or samples. In MoV, the routed units are camera-specific view features: each synchronized view of the same clip is allowed to use a different soft mixture of experts. This makes the routing explicitly view-conditioned while preserving shared expert capacity across cameras, clips, and activity domains.

Given per-view features $F=\{f_{b,v}\}$, with $f_{b,v}\in\mathbb{R}^{768}$, MoV learns a soft routing distribution over $K=12$ expert MLPs. For each view feature, a router predicts
\begin{equation}
    r_{b,v}=\operatorname{softmax}(W_r f_{b,v}+b_r), \quad r_{b,v}\in\Delta^{K-1},
\end{equation}
where $\Delta^{K-1}$ denotes the probability simplex. Each expert is an MLP $E_k:\mathbb{R}^{768}\rightarrow\mathbb{R}^{768}$ with hidden dimension 256 and GELU activation. The routed feature is computed as
\begin{equation}
    \tilde{f}_{b,v}=\sum_{k=1}^{K} r_{b,v,k} E_k(f_{b,v}).
\end{equation}

To prevent expert collapse, we add a load-balancing term that encourages the batch-averaged routing distribution $\bar{r}$ to remain close to uniform:
\begin{equation}
    \mathcal{L}_{\text{lb}} = \lambda_{\text{lb}}\, D_{\mathrm{KL}}(\bar{r}\,\Vert\,u), \qquad \lambda_{\text{lb}}=0.01,
\end{equation}
The model is trained using class-balanced cross-entropy together with this small routing regularizer.

MoV is motivated by the heterogeneous exocentric camera geometry of Ego-Exo4D. Since the third-person cameras are placed around the practitioner, each view may expose different aspects of the same skilled activity, including posture, body alignment, limb extension, footwork, and object or environment interactions. A single shared aggregation layer applies the same transformation to all views, whereas MoV allows each view to be softly routed through expert transformations that can specialize to different perspectives, without requiring camera-identity supervision. Replacing MoV with attentive pooling reduces accuracy by 6.61 percentage points, the largest single-component drop in our study (Table~\ref{tab:architecture_ablation}).

\subsection{Cross-view contrastive attention}
\label{sec:cross_view}
The routed features $\tilde{F}\in\mathbb{R}^{B\times V\times768}$ are passed to a multi-head self-attention layer that treats the $V$ views as a sequence. With 16 attention heads, each view can attend to all synchronized cameras, enabling the model to align complementary evidence before aggregation:
\begin{equation}
    A = \operatorname{MHA}(\tilde{F},\tilde{F},\tilde{F}) + \tilde{F}.
\end{equation}
A linear projection then maps the aligned features to $E\in\mathbb{R}^{B\times V\times128}$ for contrastive-style representation analysis. In auxiliary-loss experiments, these embeddings are mean-pooled and optimized with a supervised contrastive loss; however, the best configuration sets the SupCon weight to zero. The projection is retained as part of the architectural pathway, while the final objective remains class-balanced cross-entropy. Ablations reports that removing cross-view attention entirely reduces accuracy by 4.92 points (Table \ref{tab:architecture_ablation}).

\subsection{Prototype anchoring}
\label{sec:proto_anchor}
SkillMoV includes a set of four learnable prototype vectors
$P=\{p_0,p_1,p_2,p_3\}$, with $p_k\in\mathbb{R}^{768}$.
The number of prototypes matches the number of proficiency levels, but the
prototypes are optimized end-to-end and are not constrained to form an ordinal
or class-calibrated geometry.

Given the aligned per-view features
$A\in\mathbb{R}^{B\times V\times768}$, we use the prototypes as learnable
anchor tokens in a cross-attention block. The view features act as queries,
while the prototypes act as keys and values. This produces a prototype-anchored
representation, which is then pooled across views to obtain
$\bar{a}\in\mathbb{R}^{B\times768}$. We compute the cosine similarity between
$\bar{a}$ and each prototype:
\begin{equation}
    s_k =
    \frac{\bar{a}^{\top}p_k}
    {\|\bar{a}\|_2\,\|p_k\|_2},
    \qquad k\in\{0,1,2,3\}.
\end{equation}
The resulting similarity vector $s\in\mathbb{R}^{4}$ is used as an explicit
conditioning signal for the final gated projection.

\begin{figure*}[t]
    \centering
    \includegraphics[width=0.95\linewidth]{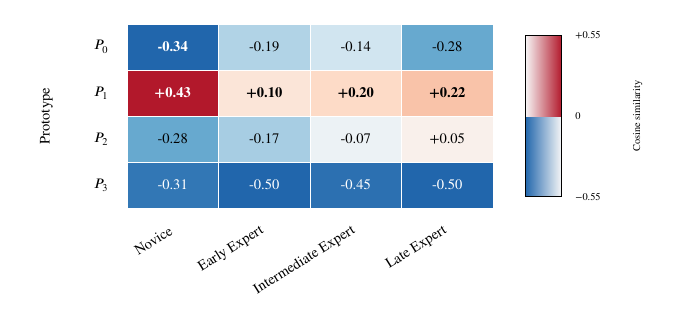}
    \caption{Mean prototype cosine similarity by ground-truth class. Rows are prototypes, columns are classes, and bold cells mark the highest-similarity prototype per class.}
    \label{fig:prototype_similarity}
\end{figure*}

This design gives the network a small set of learnable reference vectors against
which multi-view representations can be compared and reweighted. Importantly, we
do not assume that each learned prototype corresponds directly to a single
proficiency class after training. Figure~\ref{fig:prototype_similarity} reports
the mean cosine similarity between samples from each ground-truth class and the
four learned prototypes. The analysis shows that prototype $P_1$ receives the
highest average similarity across all classes, rather than producing a diagonal
or monotonic class-prototype pattern. This suggests that the prototypes are used
primarily as soft architectural anchors or query tokens, rather than as
calibrated ordinal class representatives.

Despite this limited direct interpretability, the prototype-anchoring mechanism
is useful for prediction. Removing the prototype-anchoring block reduces
accuracy by $4.07$~pp (Table~\ref{tab:architecture_ablation}), indicating that
the learned anchors provide complementary information to the final classifier.
These results suggest that prototype anchoring improves the representation, while
explicitly enforcing monotonic or class-separated prototype geometry remains an
interesting direction for future calibrated ordinal proficiency estimation.

\subsection{Prototype-conditioned gated projection}
\label{sec:proto_gate}
The prototype-anchored representation $\bar{a}$ is passed through a gated
projection module. First, an MLP expands the feature dimension:
\begin{equation}
    h = \operatorname{Dropout}(\operatorname{LN}(\operatorname{GELU}(W_h \bar{a}+b_h))),
    \quad h\in\mathbb{R}^{2560}.
\end{equation}
The gate is conditioned on both the expanded hidden features and the prototype
similarity vector:
\begin{equation}
    g = \sigma(W_g[h;s]+b_g), \qquad
    z=\operatorname{LN}(W_o(h\odot g)+b_o),
\end{equation}
where $s\in\mathbb{R}^{4}$ and $z\in\mathbb{R}^{768}$ is the final skill
embedding.

This module allows the final projection to adapt according to the relation
between the sample representation and the learned prototype anchors. Rather than
using a fixed projection for all samples, the gate reweights hidden dimensions as
a function of both the sample features and their prototype-similarity profile.
Thus, the prototypes condition the embedding pathway without requiring an
explicit ordinal loss in the best-performing configuration.

\subsection{Training objective}
The final classifier is a linear layer $W_c\in\mathbb{R}^{4\times768}$ that produces logits $\ell=W_cz+b_c$. The best configuration uses class-balanced cross-entropy:
\begin{equation}
    \mathcal{L}_{\text{CE}} = - W_y \log \frac{\exp(\ell_y)}{\sum_{j=0}^{3}\exp(\ell_j)}.
\end{equation}
We evaluated auxiliary CORAL \citep{coral}, SupCon \citep{supcon}, and prototype cross-entropy losses, but none improved the optimal Exos configuration (Table \ref{tab:loss_ablation}).

\begin{table}[t]
\centering
\caption{Scenario-level clip counts for the splits used in our experiments.}
\label{tab:dataset_breakdown}
\begin{tabular}{@{}lrrr@{}}
\toprule
Scenario & Train & Val & Test \\
\midrule
Basketball & 289 & 31 & 113 \\
Bouldering & 328 & 34 & 182 \\
Cooking & 73 & 6 & 38 \\
Dance & 223 & 26 & 167 \\
Music & 90 & 12 & 66 \\
Soccer & 36 & 6 & 24 \\
\midrule
\textbf{Total} & \textbf{1{,}039} & \textbf{115} & \textbf{590} \\
\bottomrule
\end{tabular}
\end{table}

\begin{table}[t]
\centering
\caption{Best SkillMoV training configuration. Ego, Exos, and Ego+Exos are trained and evaluated as separate runs with the same architecture and optimization recipe; Exos is the best-performing view setting.}
\label{tab:training_config}
\footnotesize
\setlength{\tabcolsep}{4pt}
\begin{tabular}{@{}ll@{}}
\toprule
Hyperparameter & Value \\
\midrule
Backbone                            & TimeSformer-base, K600 \\
LoRA rank / $\alpha$ / dropout      & 64 / 128 / 0.1 \\
Frames                              & 16 \\
View settings                       & Ego, Exos, Ego+Exos \\
Projector hidden dim.               & 2560 \\
Cross-view attention heads          & 16 \\
MoV experts / hidden dim.           & 12 / 256 \\
View dropout                        & $p=0.2$, min. 2 views; disabled for Ego \\
Epochs                              & 20 \\
Batch size / grad.\ accum.          & 16 / 4 (eff. 64) \\
Learning rate                       & $10^{-5}$, cosine, 10\% warmup \\
Optimizer                           & AdamW, wd $0.01$ \\
\bottomrule
\end{tabular}
\end{table}

\section{Experimental setup}
\label{sec:experimental_setup}

\subsection{Dataset}
\label{sec:dataset}
We evaluate SkillMoV on the EgoExo4D proficiency estimation benchmark~\citep{grauman2024egoexo4d}. Each sample contains one egocentric video and up to four synchronized exocentric views of a practitioner performing a skilled activity. The target label is a four-class proficiency level: Novice, Early Expert, Intermediate Expert, and Late Expert.

Following prior work on the same benchmark~\citep{braun2025egoppg,bianchi2025skillformer,bianchi2025pats}, we use the official training split for model development, reserving 10\% for validation, and the official held-out validation split as our test set. Table~\ref{tab:dataset_breakdown} reports the scenario-level clip counts used in our experiments.

The training split is imbalanced across proficiency levels, with $102$ Novice, $289$ Early Expert, $323$ Intermediate Expert, and $325$ Late Expert clips. We therefore use class-balanced cross-entropy, with weights computed from inverse class frequency and normalized to preserve the average loss scale:
\begin{equation}
    W = [2.0176,\; 0.7121,\; 0.6371,\; 0.6332].
\end{equation}
Removing these weights reduces overall accuracy by $2.20$ pp and harms low-frequency scenarios such as Cooking, where accuracy drops from $44.74\%$ to $10.53\%$ (Table~\ref{tab:hyper_ablation}).

\subsection{Camera configuration}
\label{sec:camera_config}
EgoExo4D provides one egocentric camera and up to four exocentric cameras. We evaluate three fixed camera configurations: Ego uses only the egocentric camera, Exos uses the four exocentric cameras, and Ego+Exos uses all five cameras. Each configuration is trained and evaluated as a separate run using the same architecture and optimization recipe, but with a different fixed input-camera set.

Stochastic view dropout follows the rule described in Section~\ref{sec:method}: it is enabled only for the multi-camera Exos and Ego+Exos runs, and disabled for Ego-only. The view ablation in Section~\ref{sec:view_ablation} shows that Exos achieves the best overall accuracy ($50.17\%$), compared with $47.63\%$ for Ego+Exos and $40.00\%$ for Ego. We use Exos as the reference configuration for component ablations, expert scaling, routing analysis, and efficiency reporting. Ego and Ego+Exos are retained in the main comparison tables to show how the same architecture behaves under different camera configurations.

\subsection{Evaluation Metrics}
\label{sec:evaluation_metrics}
Following the official Ego-Exo4D evaluation protocol for proficiency estimation, our primary metric is overall test-set accuracy. We additionally report per-scenario accuracy to expose domain-specific strengths and weaknesses across activities.

For diagnostic purposes, we also monitor an ordinal secondary metric derived from a CORAL-style cumulative head. This metric is used only for analysis and is not used for model selection or checkpoint selection. In the selected 16-frame configuration, the ordinal metric is $0.1932$; in the 32-frame ablation, it increases to $0.4271$, while categorical accuracy decreases. We discuss this temporal-resolution trade-off in Section~\ref{sec:discussion_temporal}.

\subsection{Implementation Details}
\label{sec:implementation_details}
Table~\ref{tab:training_config} summarizes the main SkillMoV training configuration. We train and evaluate the model under three fixed camera configurations: Ego, Exos, and Ego+Exos. These configurations share the same architecture and optimization recipe, but are trained as separate runs with different fixed input-camera sets. View dropout is applied only in the multi-camera Exos and Ego+Exos settings and is disabled for Ego-only.

Training uses $20$ epochs, an effective batch size of $64$, AdamW with weight decay $0.01$, and a cosine learning-rate schedule initialized at $10^{-5}$ with $10\%$ warm-up. The backbone is LoRA-adapted, gradient checkpointing is enabled, and no early stopping or test-time augmentation is used.

Frames are resized and center-cropped to $224\times224$, normalized with mean $[0.45,0.45,0.45]$ and standard deviation $[0.225,0.225,0.225]$. During training, we sample $16$ frames from a randomly positioned temporal window; at inference, we sample $16$ frames uniformly over the full clip.

\section{Results}
\label{sec:experiments}

\begin{table*}[t]
\centering
\caption{Overall top-1 accuracy (\%) on EgoExo4D proficiency estimation under Ego, Exos, and Ego+Exos views. \textbf{Best in bold}; \underline{second-best underlined}.}
\label{tab:overall}
\footnotesize
\setlength{\tabcolsep}{4pt}
\renewcommand{\arraystretch}{1.1}
\begin{tabular}{@{}lcccccccc@{}}
\toprule
Method            & Pretrain     & Ego              & Exos            & Ego+Exos         & Params   & Frames & Epochs & Type \\
\midrule
Random            & --           & 24.9             & 24.9            & 24.9             & --       & --     & --  & --             \\
Majority          & --           & 31.1             & 31.1            & 31.1             & --       & --     & --  & --             \\
\midrule
TimeSformer & --         & 42.3             & 40.1            & 40.8             & 121M     & 16     & 15  & Discriminative \\
TimeSformer & K400       & 42.9             & 39.1            & 38.6             & 121M     & 16     & 15  & Discriminative \\
TimeSformer & HowTo100M  & \underline{46.8}             & 38.2            & 39.7             & 121M     & 16     & 15  & Discriminative \\
TimeSformer & EgoVLP     & 44.4             & 40.6            & 39.5             & 121M     & 16     & 15  & Discriminative \\
TimeSformer & EgoVLPv2   & 45.9             & 38.0            & 37.8             & 121M     & 16     & 15  & Discriminative \\
EgoPulseFormer    & EgoPPG-DB    & 45.3             & 35.9            & 42.4             & 121M     & 16     & 15  & Discriminative \\
SkillFormer       & K600         & 45.9             & 46.3            & 47.5             & 14--27M  & 16--32 & 4   & Discriminative \\
SkillFormer+PATS  & K600         & \textbf{47.3} & \underline{46.6}& \underline{48.0} & 14--27M  & 24--32 & 4   & Discriminative \\
SkillMoV (ours)   & K600         & 40.0             & \textbf{50.2}   & 47.6             & 36.9M    & 16     & 20  & Discriminative \\
\midrule
ProfVLM (AGP)     & K600+SmolLM2 & 44.2             & 45.1            & \textbf{48.2}    & 5.3M     & 8      & 6   & Generative     \\
\bottomrule
\end{tabular}
\end{table*}

\subsection{Overall comparison}
\label{sec:overall_comparison}
Table~\ref{tab:overall} compares SkillMoV against the official EgoExo4D baselines and recent proficiency-estimation methods under the three standard view settings. Following prior work, we include Random and Majority-class predictors as non-learned references. We further distinguish discriminative classifiers, which predict proficiency through a dedicated classification head, from generative video-language models, which produce the label as text. The three SkillMoV results correspond to separate view-specific models trained with Ego, Exos, and Ego+Exos inputs, respectively; in each case, the model is trained jointly across all six scenarios rather than separately per scenario.

SkillMoV obtains its strongest result in the Exos setting, reaching $50.17\%$ overall accuracy. This surpasses the strongest reported Exos result among the compared methods, SkillFormer+PATS ($46.60\%$), by $+3.57$ pp. This outcome is consistent with the design of SkillMoV, which focuses on routing and aggregating complementary information across synchronized exocentric views.

In the Ego+Exos setting, SkillMoV reaches $47.63\%$, close to the best reported result in this setting, ProfVLM ($48.20\%$). The gap is $0.57$ pp, indicating that SkillMoV remains competitive when the egocentric stream is added, although its best performance is obtained from exocentric views alone. Ego-only performance is lower ($40.00\%$), consistent with a method whose main components are designed around multi-view routing and aggregation.

\begin{table*}[t]
\centering
\caption{Per-scenario top-1 accuracy (\%) by view configuration. Each row reports the breakdown of one model trained across all six scenarios, without scenario-specific tuning. In the scenario columns, \textbf{bold} and \underline{underline} mark the best and second-best results within each view block, since Ego, Exos, and Ego+Exos use different input modalities. In the \emph{Overall} column, they mark the global best and second-best results across all rows.}
\label{tab:per_scenario}
\footnotesize
\setlength{\tabcolsep}{4pt}
\renewcommand{\arraystretch}{1.05}
\begin{tabular}{@{}llrrrrrr|r@{}}
\toprule
Method & View & Basketball & Bouldering & Cooking & Dance & Music & Soccer & Overall \\
\midrule
\multirow{3}{*}{TimeSformer}
 & Ego      & 51.43              & 25.31              & \textbf{45.00}     & \textbf{55.65}     & 46.15              & 56.25              & 46.80              \\
 & Exos     & 52.30              & 17.28              & 35.00              & 42.74              & \textbf{69.23}     & 75.00              & 40.60              \\
 & Ego+Exos & 55.24              & 17.28              & 35.00              & 42.74              & 56.41              & \underline{75.00}  & 40.80              \\
\midrule
\multirow{3}{*}{EgoPulseFormer}
 & Ego      & 47.47              & 27.81              & \underline{40.00}  & \textbf{55.65}     & 46.15              & 56.25              & 46.80              \\
 & Exos     & 49.29              & 12.58              & 40.00              & 42.74              & \textbf{69.23}     & 75.00              & 40.60              \\
 & Ego+Exos & 52.52              & 18.54              & 40.00              & 42.74              & 56.41              & \underline{75.00}  & 40.80              \\
\midrule
\multirow{3}{*}{SkillFormer}
 & Ego      & \textbf{69.03}     & 30.77              & 31.58              & 20.51              & \textbf{72.41}     & \underline{70.83}  & 45.90              \\
 & Exos     & \textbf{70.80}     & 33.52              & 47.37              & 15.38              & \underline{68.97}  & 66.67              & 46.30              \\
 & Ego+Exos & \textbf{77.88}     & 31.87              & \textbf{60.53}     & 13.68              & \underline{68.10}  & 66.67              & 47.50              \\
\midrule
\multirow{3}{*}{SkillFormer+PATS}
 & Ego      & \underline{64.60}  & \textbf{42.31}     & 21.05              & 19.66              & 70.69              & 66.67              & 47.29              \\
 & Exos     & \underline{68.14}  & 31.32              & \textbf{60.53}     & 20.51              & 67.24              & 66.67              & 46.61              \\
 & Ego+Exos & \underline{75.22}  & 31.32              & 42.11              & 24.79              & \textbf{68.97}     & 66.67              & 48.00              \\
\midrule
\multirow{3}{*}{ProfVLM (AGP)}
 & Ego      & 36.00              & \underline{37.48}  & 31.00              & \underline{51.41}  & \underline{72.05}  & 57.25              & 44.20              \\
 & Exos     & 33.00              & \textbf{37.48}     & \underline{56.00}  & \textbf{53.85}     & 61.53              & \underline{76.00}  & 45.10              \\
 & Ego+Exos & 41.00              & \textbf{38.74}     & \underline{51.00}  & \textbf{60.35}     & 56.26              & 69.75              & \underline{48.20}  \\
\midrule
\multirow{3}{*}{SkillMoV (ours)}
 & Ego      & \underline{64.60}  & 19.23              & 26.32              & 32.34              & 68.18              & \textbf{79.17}     & 40.00              \\
 & Exos     & 64.60              & \underline{36.81}  & 44.74              & \underline{46.11}  & 65.15              & \textbf{79.17}     & \textbf{50.17}     \\
 & Ego+Exos & 67.26              & \underline{33.52}  & 36.84              & \underline{43.11}  & 59.09              & \textbf{79.17}     & 47.63              \\
\bottomrule
\end{tabular}
\end{table*}

\begin{table}[t]
\centering
\caption{Summary of Exos per-scenario robustness. \emph{Min} and \emph{Std.} are computed across the six scenario accuracies in the Exos row. \emph{Avg. rank} is computed by ranking methods within each Exos scenario column.}
\label{tab:exos_robustness}
\footnotesize
\setlength{\tabcolsep}{5pt}
\renewcommand{\arraystretch}{1.05}
\begin{tabular}{@{}lcccc@{}}
\toprule
Method & Overall $\uparrow$ & Min $\uparrow$ & Std. $\downarrow$ & Avg. rank $\downarrow$ \\
\midrule
TimeSformer        & 40.60 & 17.28 & 19.73 & 3.67 \\
EgoPulseFormer     & 40.60 & 12.58 & 20.52 & 3.83 \\
SkillFormer        & 46.30 & 15.38 & 20.60 & 3.50 \\
SkillFormer+PATS   & 46.61 & 20.51 & 19.14 & 3.50 \\
ProfVLM (AGP)      & 45.10 & 33.00 & \textbf{14.45} & 3.00 \\
SkillMoV (ours)    & \textbf{50.17} & \textbf{36.81} & 14.65 & \textbf{2.83} \\
\bottomrule
\end{tabular}
\end{table}

\subsection{Per-scenario comparison}
\label{sec:per_scenario_comparison}
Table~\ref{tab:per_scenario} reports scenario-level accuracy for SkillMoV and compares it with the official EgoExo4D TimeSformer baselines and the strongest available prior results. This breakdown is important because proficiency estimation is highly heterogeneous across activities: some scenarios benefit more from egocentric evidence or language supervision, whereas others are better captured by exocentric discriminative models.

All per-scenario values follow the same evaluation protocol. For each method and view configuration, the six scenario columns report the test-set breakdown of one model trained jointly across all six EgoExo4D scenarios. The table therefore evaluates how well each unified model generalizes across activities under a fixed camera configuration.

The per-scenario winners are distributed across methods, showing that no single model dominates every activity. SkillFormer Ego+Exos obtains the best result on Basketball ($77.88\%$) and ties the best result on Cooking ($60.53\%$), SkillFormer Ego performs best on Music ($72.41\%$), SkillFormer+PATS Ego performs best on Bouldering ($42.31\%$), ProfVLM Ego+Exos performs best on Dance ($60.35\%$), and SkillMoV obtains the best result on Soccer across all three view configurations ($79.17\%$).

Table~\ref{tab:exos_robustness} summarizes the Exos setting, where SkillMoV obtains its strongest overall result. In this configuration, SkillMoV achieves the highest overall accuracy ($50.17\%$), the highest per-scenario floor among the compared Exos models ($36.81\%$), and the best average per-scenario rank ($2.83$). Its dispersion across scenarios is also low ($14.65$ pp), close to ProfVLM Exos ($14.45$ pp), while maintaining a higher overall accuracy. This indicates that SkillMoV's Exos performance is not driven only by a single scenario peak, but by a comparatively balanced profile across activities.

Compared with the official TimeSformer Exos baseline, SkillMoV improves on five of six scenarios: Basketball ($+12.30$ pp), Bouldering ($+19.53$ pp), Cooking ($+9.74$ pp), Dance ($+3.37$ pp), and Soccer ($+4.17$ pp), with a decrease on Music ($-4.08$ pp). The largest gains occur in scenarios where the TimeSformer baseline is weakest, suggesting that routing and prototype-conditioned projection are most useful when a single shared representation struggles to cover heterogeneous visual evidence.

Overall, the per-scenario analysis shows that SkillMoV does not dominate every activity, but its Exos configuration achieves the best overall accuracy and avoids the low per-scenario scores observed in several competing Exos models.

\section{Ablation Studies}
\label{sec:ablations}

\begin{table}[t]
\centering
\caption{Architecture ablation on the selected Exos configuration. $\Delta$ is the drop relative to the full model.}
\label{tab:architecture_ablation}
\footnotesize
\setlength{\tabcolsep}{4pt}
\begin{tabular}{@{}lrr@{}}
\toprule
Configuration                                            & Accuracy        & $\Delta$ \\
\midrule
SkillMoV (full model)                                    & \textbf{50.17}  & --      \\
-- MoV router (attentive pooling instead)                & 43.56           & -6.61   \\
-- cross-view attention                                  & 45.25           & -4.92   \\
-- prototype anchoring (mean-pool + linear head)          & 46.10           & -4.07   \\
-- stochastic view dropout                               & 46.27           & -3.90   \\
\bottomrule
\end{tabular}
\end{table}

\subsection{Architecture ablation}
\label{sec:architecture_ablation}

Table~\ref{tab:architecture_ablation} isolates the contribution of the main architectural components. All variants use the same Exos input, training schedule, and class-balanced cross-entropy objective.

The MoV router is the largest contributor: replacing expert routing with attentive pooling reduces accuracy by $6.61$~pp. This indicates that applying a shared aggregation mechanism to all camera features is less effective than allowing view features to be routed through multiple expert transformations. Removing cross-view attention causes the second-largest drop ($4.92$~pp), suggesting that synchronized cameras benefit from interacting before the final aggregation stage. Prototype anchoring and stochastic view dropout also contribute meaningful gains, with drops of $4.07$~pp and $3.90$~pp respectively, showing that both learnable anchors and partial-view regularization improve the final representation.

\begin{table*}[t]
\centering
\caption{View ablation on SkillMoV. $\Delta$ is the overall-accuracy change relative to Exos, the best view setting.}
\label{tab:view_ablation}
\footnotesize
\setlength{\tabcolsep}{4pt}
\begin{tabular}{@{}llrrrrrr|rr@{}}
\toprule
Configuration           & Cameras        & Basketball     & Bouldering     & Cooking        & Dance          & Music          & Soccer         & Overall        & $\Delta$    \\
\midrule
Ego                     & 1 ego          & 64.60          & 19.23          & 26.32          & 32.34          & 68.18          & 79.17          & 40.00          & $-10.17$    \\
Ego+Exos                & 1 ego + 4 exos & 67.26          & 33.52          & 36.84          & 43.11          & 59.09          & 79.17          & 47.63          &  $-2.54$    \\
\textbf{Exos}           & \textbf{4 exos} & \textbf{64.60} & \textbf{36.81} & \textbf{44.74} & \textbf{46.11} & \textbf{65.15} & \textbf{79.17} & \textbf{50.17} & \textbf{---} \\
PATS-style sampling     & Ego+Exos       & 60.18          & 41.21          & 36.84          & 44.31          & 63.64          & 83.33          & 49.66          &  $-0.51$    \\
\bottomrule
\end{tabular}
\end{table*}

\subsection{View and camera study}
\label{sec:view_ablation}
Table~\ref{tab:view_ablation} compares Ego, Exos, Ego+Exos, and a PATS-style temporal sampling variant. Exos achieves the best overall accuracy ($50.17\%$), while Ego+Exos reaches $47.63\%$ and Ego-only reaches $40.00\%$. This pattern suggests that the current SkillMoV design extracts more reliable proficiency cues from exocentric views than from the egocentric or mixed streams.

The per-scenario breakdown shows that the egocentric view is not uniformly beneficial. Adding Ego to Exos reduces overall accuracy by $2.54$~pp, with drops on Bouldering, Cooking, Dance, and Music. These scenarios often require observing the practitioner's full-body configuration or the relation between body motion and the surrounding environment, which can be partially occluded or absent in the first-person view. The PATS-style sampling variant recovers much of the gap to Exos, reaching $49.66\%$, and improves over Exos on Bouldering and Soccer. However, it degrades Basketball and Cooking, leading to a slightly lower overall score. These results suggest that view selection and temporal sampling interact strongly, and that adaptive temporal strategies may be most useful when paired with the appropriate camera configuration.

\begin{table}[t]
\centering
\caption{Loss-function ablation. Ego+Exos uses CE-only as reference; the final row tests the best auxiliary loss on Exos.}
\label{tab:loss_ablation}
\footnotesize
\setlength{\tabcolsep}{4pt}
\begin{tabular}{@{}lrr@{}}
\toprule
Configuration                                        & Accuracy        & $\Delta$ vs. CE      \\
\midrule
CE-only (Ego+Exos)                                   & \textbf{47.63}  & --                  \\
+ ProtoCE, $\lambda=0.10$                            & 48.14           & +0.51               \\
+ ProtoCE, $\lambda=0.05$                            & 46.95           & -0.68               \\
+ ProtoCE, $\lambda=0.30$                            & 45.42           & -2.21               \\
+ CORAL ordinal                                      & 45.25           & -2.38               \\
+ SupCon contrastive                                 & 44.41           & -3.22               \\
\midrule
Exos + ProtoCE, $\lambda=0.10$                       & 46.78           & -3.39 vs.\ selected \\
\bottomrule
\end{tabular}
\end{table}

\subsection{Loss-function ablation}
\label{sec:loss_ablation}

Table~\ref{tab:loss_ablation} reports auxiliary-loss experiments used as diagnostics. Most variants are evaluated in the Ego+Exos setting, where auxiliary prototype alignment was initially explored; the final row tests the best auxiliary loss on the selected Exos configuration.

Class-balanced cross-entropy remains the most stable objective. Prototype cross-entropy with $\lambda=0.10$ gives a small gain in Ego+Exos, but the same auxiliary loss degrades the selected Exos configuration. CORAL and supervised contrastive learning also reduce accuracy. These results indicate that, for the current dataset size and class/domain imbalance, additional auxiliary losses do not consistently improve the final classifier. We therefore adopt class-balanced cross-entropy alone for the selected Exos model.

\begin{table}[t]
\centering
\caption{Training and hyperparameter ablation. $\Delta$ is relative to the best SkillMoV configuration.}
\label{tab:hyper_ablation}
\footnotesize
\setlength{\tabcolsep}{4pt}
\begin{tabular}{@{}lrr@{}}
\toprule
Configuration                                  & Accuracy        & $\Delta$ \\
\midrule
SkillMoV (selected, 16 frames, $r{=}64$)       & \textbf{50.17}  & --      \\
32 frames                                      & 45.76           & -4.41   \\
LR $5\times10^{-6}$, 30 epochs                 & 46.10           & -4.07   \\
Label smoothing 0.1                            & 46.78           & -3.39   \\
LoRA $r=32$, $\alpha=64$                       & 47.97           & -2.20   \\
Uniform CE, no class weights                   & 47.97           & -2.20   \\
\bottomrule
\end{tabular}
\end{table}

\subsection{Training and hyperparameter ablation}
\label{sec:hyper_ablation}

Table~\ref{tab:hyper_ablation} summarizes the main training and hyperparameter ablations. The selected configuration uses 16 frames, LoRA rank $64$, and class-balanced cross-entropy.

Increasing the number of frames from 16 to 32 reduces categorical accuracy by $4.41$~pp, despite improving the ordinal secondary metric discussed in Section~\ref{sec:discussion_temporal}. A lower learning rate with longer training also underperforms, suggesting that the selected 20-epoch schedule provides sufficient adaptation. Halving the LoRA rank reduces accuracy by $2.20$~pp, indicating that rank $64$ provides a better capacity-efficiency trade-off. Removing class weights produces the same drop, confirming the importance of class-balanced training for this imbalanced proficiency-estimation setting.

\begin{table}[t]
\centering
\caption{MoV expert-scaling study on the multi-camera settings.}
\label{tab:expert_scaling}
\footnotesize
\setlength{\tabcolsep}{5pt}
\begin{tabular}{@{}lrr@{}}
\toprule
Experts & Exos & Ego+Exos \\
\midrule
$K=4$  & 48.10          & 47.29          \\
$K=8$  & 47.63          & 46.44          \\
$K=12$ & \textbf{50.17} & \textbf{47.63} \\
\bottomrule
\end{tabular}
\end{table}

\subsection{Mixture-of-View expert scaling}
\label{sec:expert_scaling}

Table~\ref{tab:expert_scaling} studies the number of MoV experts in the two multi-camera settings. Ego is omitted because it contains only a single camera, while the expert-scaling question is primarily about multi-view routing capacity.

The $K=12$ configuration performs best in both Exos and Ego+Exos. Smaller expert counts underperform, suggesting that the synchronized camera set benefits from additional routing capacity across views and scenarios. Since Exos is also the strongest view configuration overall, we use $K=12$ for the selected model and all component ablations.

\subsection{Mixture-of-View routing analysis}
\label{sec:mov_analysis}

To complement the architectural ablation, we analyze the routing distribution learned by the MoV router. This analysis is performed on the selected Exos model. We extract the per-view routing weights $r_{b,v}\in\Delta^{K-1}$ from all clips in the official EgoExo4D test set, corresponding to $590$ clips and four exocentric views, for a total of $2{,}360$ view instances.

\begin{figure*}[t]
    \centering
    \includegraphics[width=0.95\textwidth]{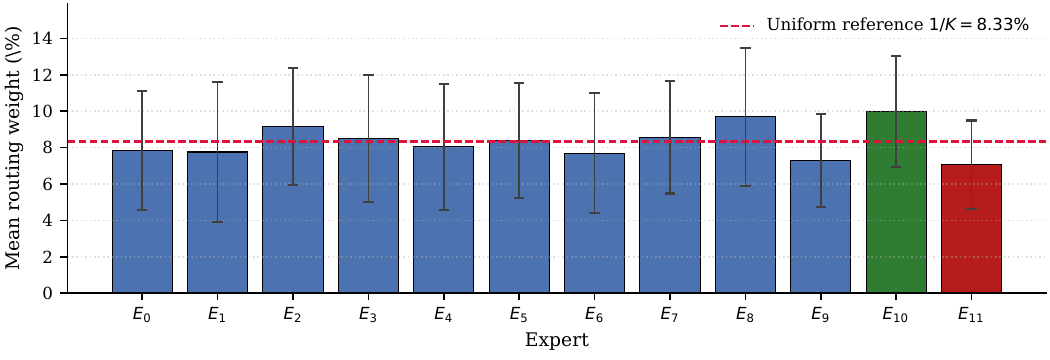}
    \caption{MoV expert utilization on the EgoExo4D test set. Bars show mean routing weight with one-standard-deviation error bars; the dashed line marks uniform routing.}
    \label{fig:mov_utilization}
\end{figure*}

\paragraph{Expert utilization.}
Figure~\ref{fig:mov_utilization} reports the mean routing weight assigned to each of the $K=12$ experts. The utilization values remain close to the uniform reference $1/K\approx8.33\%$, with no expert receiving negligible mass. This indicates that the load-balancing term prevents degenerate expert collapse and keeps the mixture broadly active across the test set.

\begin{figure*}[t]
    \centering
    \includegraphics[width=0.95\textwidth]{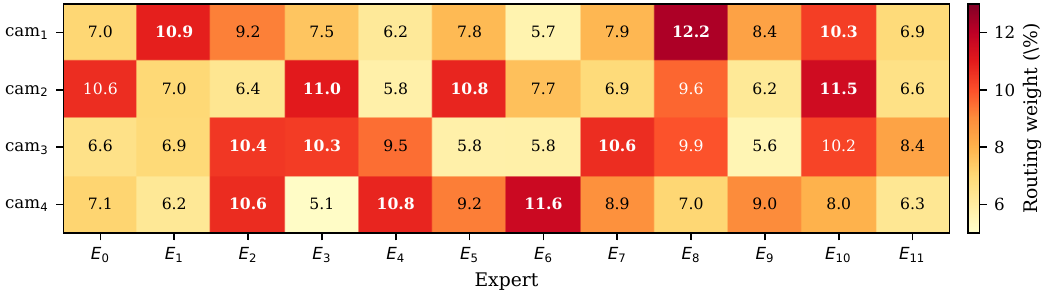}
    \caption{MoV view-wise routing for the selected Exos model. Rows are exocentric cameras; cell values show mean routing weights in percent.}
    \label{fig:mov_view_routing}
\end{figure*}

\paragraph{View-wise routing.}
Figure~\ref{fig:mov_view_routing} reports the mean routing distribution conditioned on camera index. The four exocentric cameras exhibit different routing preferences: for example, cam$_1$ assigns relatively high mass to $E_8$, $E_1$, and $E_{10}$, whereas cam$_4$ emphasizes $E_6$, $E_4$, and $E_2$. However, the pattern is not a hard camera-to-expert assignment. Several experts are shared across cameras, and each camera retains a distributed routing profile. This suggests that MoV learns soft view-dependent expert preferences without requiring camera-identity supervision.

\begin{figure*}[t]
    \centering
    \includegraphics[width=0.95\textwidth]{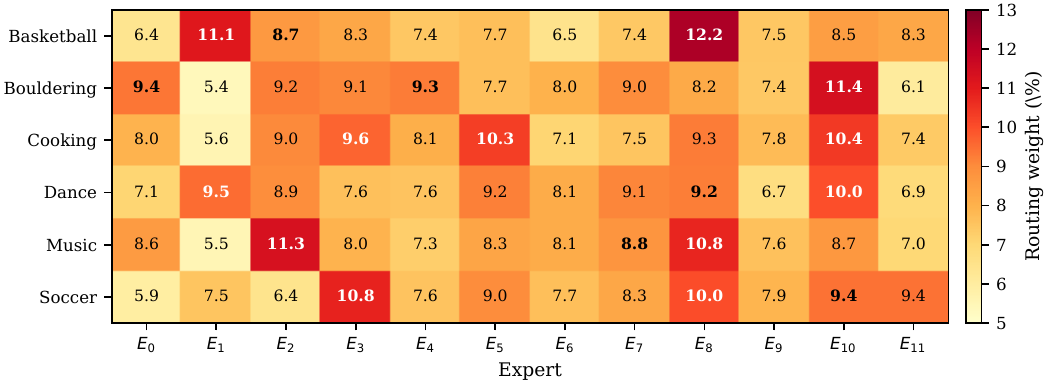}
    \caption{MoV scenario-wise routing for the selected Exos model. Rows are activity domains; cell values show mean routing weights in percent.}
    \label{fig:mov_scenario_routing}
\end{figure*}

\paragraph{Scenario-wise routing.}
Figure~\ref{fig:mov_scenario_routing} aggregates routing distributions by activity domain. The routing profile also varies across scenarios. Basketball emphasizes $E_8$ and $E_1$, Music emphasizes $E_2$ and $E_8$, and Soccer emphasizes $E_3$ and $E_8$, while Bouldering, Cooking, and Dance show more distributed profiles. As in the view-wise analysis, these are soft preferences rather than discrete scenario-to-expert assignments. The router is not given scenario labels, so these differences emerge from the proficiency objective and the visual structure of the input clips.

Overall, the routing analysis supports the intended role of MoV as a soft view-conditioned mixture. Expert utilization remains close to uniform, while the view-wise and scenario-wise heatmaps show non-identical routing preferences across cameras and activities. Importantly, these patterns remain distributed: MoV does not collapse to a few experts and does not impose fixed camera-to-expert or scenario-to-expert mappings. Instead, it learns soft expert preferences that vary with viewpoint and activity domain.

\begin{table*}[t]
\centering
\caption{Measured efficiency of the selected Exos configuration. Inference memory is measured in fp16 with batch size 1 on one NVIDIA H200.}
\label{tab:efficiency}
\footnotesize
\setlength{\tabcolsep}{5pt}
\begin{tabular}{@{}lrrrrr@{}}
\toprule
Model & Total params & Trainable & Trainable (\%) & Train time & Infer. mem. \\
\midrule
TimeSformer + LoRA $r=64$ & 148.3M & 27.0M & 18.22 & 04:24:02 & 0.79 GB \\
SkillMoV (selected Exos)  & 158.1M & 36.9M & 23.32 & 04:27:49 & 0.81 GB \\
\bottomrule
\end{tabular}
\end{table*}

\subsection{Computational efficiency}
\label{sec:efficiency}

Table~\ref{tab:efficiency} reports the measured cost of the selected Exos model relative to the LoRA-adapted TimeSformer baseline under the same evaluation setup. SkillMoV adds approximately $10$M total parameters and $9.9$M trainable parameters, increasing the trainable fraction from $18.22\%$ to $23.32\%$.

Despite the larger projector, the measured overhead is small in our setup: training time increases from 04:24:02 to 04:27:49, and fp16 inference memory increases from $0.79$ GB to $0.81$ GB. This suggests that the proposed MoV projector adds limited overhead relative to the LoRA-adapted multi-view backbone.

\section{Discussion}
\label{sec:discussion}

\subsection{Why exocentric views dominate}
\label{sec:discussion_exos}

The strongest empirical pattern in our study is the advantage of the Exos configuration. This is consistent with the way proficiency labels are collected in EgoExo4D: human annotators judge skill from an external perspective, and exocentric cameras capture similar evidence. In contrast, egocentric video primarily records what the practitioner sees, which may omit global posture, lower-body coordination, body-object relations, or body-environment interactions.

This helps explain why adding the egocentric stream does not necessarily improve performance. Ego+Exos contains more visual input than Exos, but its overall accuracy is lower by $2.54$~pp. The result suggests that more views are not automatically better; the usefulness of a view depends on how well it matches the visual evidence underlying the proficiency labels.

\subsection{The temporal-resolution trade-off}
\label{sec:discussion_temporal}

Increasing the number of frames from 16 to 32 reduces categorical accuracy from $50.17\%$ to $45.76\%$, while increasing the ordinal secondary metric from $0.1932$ to $0.4271$. This suggests a trade-off between categorical discrimination and ordinal structure. Longer temporal coverage may help the model recover progression cues that are useful for ordering proficiency levels, but uniform 32-frame sampling may also dilute the salient moments that drive categorical decisions.

This observation should be read together with the PATS-style sampling result in Table~\ref{tab:view_ablation}. On Ego+Exos, PATS-style sampling recovers much of the gap to the selected Exos configuration, reaching $49.66\%$, but it remains slightly below Exos and shows mixed per-scenario effects. This suggests that temporal selection is useful, but its benefit depends on the camera configuration and activity domain. Future work should therefore focus on adaptive temporal strategies that are jointly view-aware and scenario-sensitive, rather than simply increasing the number of uniformly sampled frames.

\subsection{MoV as soft view-dependent routing}
\label{sec:discussion_mov}

The ablation results and routing analysis both support the role of MoV as a view-dependent aggregation mechanism. Replacing MoV with attentive pooling reduces accuracy by $6.61$~pp, the largest component drop in Table~\ref{tab:architecture_ablation}. This suggests that routing view features through multiple expert transformations is more effective than passing all views through a single shared aggregation module.

The routing figures clarify how this mechanism behaves. Expert utilization remains close to uniform, indicating that the load-balancing term prevents collapse to a small subset of experts. At the same time, the view-wise and scenario-wise heatmaps show different routing preferences across cameras and activities. These preferences are not hard assignments: several experts are shared across cameras and scenarios, and the routing distributions remain distributed. Thus, MoV is best interpreted as a soft routing mechanism that modulates expert usage according to the input view and activity, rather than as a fixed camera-to-expert mapping.

\subsection{Prototype behaviour}
\label{sec:discussion_ordinal}

The prototype similarity analysis in Figure~\ref{fig:prototype_similarity} shows that the learned prototypes should not be interpreted as a calibrated ordinal scale. Although one prototype is learned per proficiency level, prototype $P_1$ receives the highest average cosine similarity across all ground-truth classes. Thus, the learned geometry is not monotonic along the skill axis.

This does not mean that prototype anchoring is ineffective. Removing the prototype-anchoring block reduces accuracy by $4.07$~pp, indicating that the prototypes improve the discriminative embedding. A more precise interpretation is that the prototypes act as learnable soft anchors or query tokens that condition the final representation, rather than as directly interpretable ordinal rulers. Enforcing monotonic inter-prototype geometry remains a promising extension for calibrated ordinal proficiency estimation.

\subsection{Per-scenario behaviour}
\label{sec:discussion_per_scenario}

The per-scenario results indicate that SkillMoV benefits most from exocentric views in activities where externally visible body motion provides strong proficiency cues. Soccer is the clearest case: SkillMoV reaches $79.17\%$ in all three view configurations, although this result should be interpreted cautiously because the Soccer test subset contains only $24$ clips. Basketball is also strong in the Exos setting ($64.60\%$), suggesting that repeated and spatially structured motion patterns are well captured by the proposed multi-view representation.

Other activities remain more challenging. Bouldering improves substantially from Ego-only to Exos ($19.23\%$ to $36.81\%$), but remains the weakest Exos scenario, likely because climbing proficiency depends on route-specific body-wall interactions and highly variable movement patterns. Cooking and Dance also show remaining gaps to the best per-scenario results achieved by other methods, suggesting that they may require more precise temporal localization, hand-object reasoning, or pose-sensitive motion cues.

Overall, the per-scenario results suggest that the remaining errors are not explained by a single failure mode. SkillMoV improves the overall Exos profile, but closing the remaining domain-specific gaps likely requires temporal selection and visual cues tailored to the structure of each activity.

\section{Limitations and future work}
\label{sec:limitations}

Several limitations remain. First, although prototype anchoring improves accuracy, the learned prototypes should not be interpreted as calibrated ordinal anchors. Their cosine-similarity structure is not monotonic across proficiency levels, so they are better viewed as learnable conditioning anchors than as directly interpretable ordinal reference points. Explicit monotonic regularization would be required before using them for calibrated ordinal feedback.

Second, our per-scenario analysis is based on aggregate accuracies. Confusion matrices and qualitative inspection of misclassified clips would help distinguish adjacent-level ordinal errors from domain-specific recognition failures. This is especially relevant for scenarios such as Bouldering, Cooking, and Dance, where residual errors remain substantial.

Third, some scenario-specific test subsets are small. For example, Soccer contains only 24 clips, so a single prediction changes accuracy by $4.17$~pp. We therefore interpret per-scenario peaks cautiously and emphasize overall accuracy and cross-scenario trends rather than isolated per-scenario margins.

Finally, our efficiency analysis reports measured training time, trainable parameters, and inference memory, but does not include a full FLOPs/GMACs comparison for the LoRA-adapted variants. Accurate operation-level profiling is non-trivial because the PEFT LoRA wrappers used in our implementation are not directly supported by the profiling tool we used. The efficiency results should therefore be interpreted as measured system-level costs rather than as a complete operation-level analysis.

Future work will address these limitations in four directions. First, we will investigate explicit monotonic constraints on the prototype geometry, so that the learned anchors become not only discriminative but also ordinally calibrated. Second, we will extend the evaluation with per-scenario confusion matrices and qualitative failure analysis, with particular attention to whether errors occur mostly between adjacent proficiency levels or across distant classes. Third, adaptive temporal sampling remains a promising direction: the current results suggest that longer temporal coverage can improve ordinal cues, but may dilute the key moments needed for categorical classification. Fourth, knowledge distillation could be used to transfer the behaviour of larger or multi-view teacher models into lighter student architectures, reducing deployment cost while preserving the benefits of MoV routing and prototype-conditioned gating. Teacher--student distillation approaches ~\citep{meneghetti1, meneghetti2} provide natural starting points for compressing SkillMoV into efficient models suitable for real-time or resource-constrained skill-assessment systems.

\section{Conclusion}
\label{sec:conclusion}

We presented SkillMoV, a parameter-efficient framework for multi-view proficiency estimation from synchronized EgoExo4D video. SkillMoV combines LoRA-adapted TimeSformer features with a Mixture-of-View projector composed of soft expert routing, cross-view attention, prototype anchoring, and prototype-conditioned gated projection.

On EgoExo4D proficiency estimation, SkillMoV reaches $50.17\%$ overall accuracy in the Exos configuration, surpassing the strongest reported Exos result among the compared methods by $3.57$~pp. In Ego+Exos, it remains competitive with the best reported result in that setting ($47.63\%$ versus $48.20\%$). Beyond the aggregate score, the strength of SkillMoV lies in its balanced multi-view behaviour: it uses view-conditioned expert routing rather than a single shared projection, improves the Exos setting without scenario-specific training, and preserves soft expert sharing instead of collapsing to fixed camera-specific assignments.

Ablations confirm the contribution of each component. MoV routing is the largest contributor, followed by cross-view attention, prototype anchoring, and stochastic view dropout. The routing analysis further shows that all experts remain active and that routing preferences vary across cameras and scenarios without explicit camera or scenario supervision. These findings support the central hypothesis of SkillMoV: proficiency estimation benefits from multi-view representations that are view-conditioned, softly shared, and not forced through a single shared projection.

Future work should focus on three directions: calibrated ordinal prototype geometry, adaptive temporal sampling, and detailed failure analysis with confusion matrices and qualitative inspection of misclassified clips. These extensions could improve both the accuracy and interpretability of multi-view proficiency estimation systems.

\section*{Declaration of competing interest}
The authors declare that they have no known competing financial interests or personal relationships that could have appeared to influence the work reported in this paper.

\section*{Data availability}
This study uses the EgoExo4D benchmark, which is publicly available from the dataset providers subject to their license and access terms. Experimental scripts and model checkpoints will be released upon acceptance, subject to institutional and dataset-distribution constraints.

\printcredits

\bibliographystyle{cas-model2-names}

%% Loading bibliography database
\bibliography{skillmov}

@inproceedings{meneghetti1,
	address = {Cham},
	author = {Meneghetti, Laura and Bianchi, Edoardo and Demo, Nicola and Rozza, Gianluigi},
	booktitle = {Design and Architecture for Signal and Image Processing},
	editor = {Lorandel, Jordane and Kamaleldin, Ahmed},
	isbn = {978-3-031-87897-8},
	pages = {81--92},
	publisher = {Springer Nature Switzerland},
	title = {KD-AHOSVD: Neural Network Compression via Knowledge Distillation and Tensor Decomposition},
	year = {2025}}

@article{meneghetti2,
	author = {Laura Meneghetti and Edoardo Bianchi and Nicola Demo and Gianluigi Rozza},
	doi = {https://doi.org/10.1016/j.sysarc.2026.103778},
	issn = {1383-7621},
	journal = {Journal of Systems Architecture},
	pages = {103778},
	title = {Plug-and-play neural compression: A knowledge distillation framework with flexible dimensionality reduction},
	url = {https://www.sciencedirect.com/science/article/pii/S1383762126000962},
	volume = {175},
	year = {2026}}

@misc{coral,
      title={CORAL: Contextual Response Retrievability Loss Function for Training Dialog Generation Models}, 
      author={Bishal Santra and Ravi Ghadia and Manish Gupta and Pawan Goyal},
      year={2023},
      eprint={2205.10558},
      archivePrefix={arXiv},
      primaryClass={cs.CL},
      url={https://arxiv.org/abs/2205.10558}, 
}

@inproceedings{supcon,
	author = {Khosla, Prannay and Teterwak, Piotr and Wang, Chen and Sarna, Aaron and Tian, Yonglong and Isola, Phillip and Maschinot, Aaron and Liu, Ce and Krishnan, Dilip},
	booktitle = {Advances in Neural Information Processing Systems},
	editor = {H. Larochelle and M. Ranzato and R. Hadsell and M.F. Balcan and H. Lin},
	pages = {18661--18673},
	publisher = {Curran Associates, Inc.},
	title = {Supervised Contrastive Learning},
	url = {https://proceedings.neurips.cc/paper_files/paper/2020/file/d89a66c7c80a29b1bdbab0f2a1a94af8-Paper.pdf},
	volume = {33},
	year = {2020}}

@inproceedings{bertasius2021timesformer,
  author    = {Bertasius, G. and Wang, H. and Torresani, L.},
  title     = {Is space-time attention all you need for video understanding?},
  booktitle = {Proceedings of the International Conference on Machine Learning},
  year      = {2021},
  pages     = {813--824}
}

@inproceedings{bianchi2025pats,
  author    = {Bianchi, E. and Liotta, A.},
  title     = {{PATS}: Proficiency-aware temporal sampling for multi-view sports skill assessment},
  booktitle = {2025 IEEE International Workshop on Sport, Technology and Research (STAR)},
  year      = {2025},
  pages     = {1--6},
  doi       = {10.1109/STAR66750.2025.11264769}
}

@article{bianchi2025skillformer,
  author  = {Bianchi, E. and Liotta, A.},
  title   = {{SkillFormer}: Unified multi-view video understanding for proficiency estimation},
  journal = {arXiv preprint arXiv:2505.08665},
  year    = {2025}
}

@article{bianchi2026profvlm,
  author  = {Bianchi, E. and Staiano, J. and Liotta, A.},
  title   = {{ProfVLM}: A lightweight video-language model for multi-view proficiency estimation},
  journal = {Computer Vision and Image Understanding},
  volume  = {268},
  year    = {2026},
  pages   = {104749}
}

@article{braun2025egoppg,
  author  = {Braun, B. and Armani, R. and Meier, M. and Moebus, M. and Holz, C.},
  title   = {{egoPPG}: Heart rate estimation from eye-tracking cameras in egocentric systems to benefit downstream vision tasks},
  journal = {arXiv preprint arXiv:2502.20879},
  year    = {2025}
}

@article{cao2020coral,
  author  = {Cao, W. and Mirjalili, V. and Raschka, S.},
  title   = {Rank consistent ordinal regression for neural networks with application to age estimation},
  journal = {Pattern Recognition Letters},
  volume  = {140},
  year    = {2020},
  pages   = {325--331}
}

@article{fedus2022switch,
  author  = {Fedus, W. and Zoph, B. and Shazeer, N.},
  title   = {Switch transformers: Scaling to trillion parameter models with simple and efficient sparsity},
  journal = {Journal of Machine Learning Research},
  volume  = {23},
  year    = {2022},
  pages   = {1--39}
}

@inproceedings{grauman2024egoexo4d,
  author    = {Grauman, K. and Westbury, A. and Torresani, L. and Kitani, K. and Malik, J. and others},
  title     = {{Ego-Exo4D}: Understanding skilled human activity from first- and third-person perspectives},
  booktitle = {Proceedings of the IEEE/CVF Conference on Computer Vision and Pattern Recognition},
  year      = {2024},
  pages     = {19383--19400}
}

@inproceedings{hu2022lora,
  author    = {Hu, E. J. and Shen, Y. and Wallis, P. and Allen-Zhu, Z. and Li, Y. and Wang, S. and Wang, L. and Chen, W.},
  title     = {{LoRA}: Low-rank adaptation of large language models},
  booktitle = {Proceedings of the International Conference on Learning Representations},
  year      = {2022},
  pages     = {1--13}
}

@inproceedings{khosla2020supervised,
  author    = {Khosla, P. and Teterwak, P. and Wang, C. and Sarna, A. and Tian, Y. and Isola, P. and Maschinot, A. and Liu, C. and Krishnan, D.},
  title     = {Supervised contrastive learning},
  booktitle = {Advances in Neural Information Processing Systems},
  year      = {2020},
  pages     = {18661--18673}
}

@inproceedings{liu2022videoswin,
  author    = {Liu, Z. and Ning, J. and Cao, Y. and Wei, Y. and Zhang, Z. and Lin, S. and Hu, H.},
  title     = {Video {Swin} Transformer},
  booktitle = {Proceedings of the IEEE/CVF Conference on Computer Vision and Pattern Recognition},
  year      = {2022},
  pages     = {3202--3211}
}

@inproceedings{shazeer2017outrageously,
  author    = {Shazeer, N. and Mirhoseini, A. and Maziarz, K. and Davis, A. and Le, Q. and Hinton, G. and Dean, J.},
  title     = {Outrageously large neural networks: The sparsely-gated mixture-of-experts layer},
  booktitle = {Proceedings of the International Conference on Learning Representations},
  year      = {2017},
  pages     = {1--19}
}

@inproceedings{snell2017prototypical,
  author    = {Snell, J. and Swersky, K. and Zemel, R.},
  title     = {Prototypical networks for few-shot learning},
  booktitle = {Advances in Neural Information Processing Systems},
  year      = {2017},
  pages     = {4077--4087}
}

@article{zhou2024aqasurvey,
  author  = {Zhou, K. and Cai, R. and Wang, L. and Shum, H. P. H. and Liang, X.},
  title   = {A comprehensive survey of action quality assessment: Method and benchmark},
  journal = {arXiv preprint arXiv:2412.11149},
  year    = {2024}
}

@inproceedings{tran2015c3d,
  author    = {Tran, D. and Bourdev, L. and Fergus, R. and Torresani, L. and Paluri, M.},
  title     = {Learning spatiotemporal features with 3{D} convolutional networks},
  booktitle = {Proceedings of the IEEE International Conference on Computer Vision},
  year      = {2015},
  pages     = {4489--4497}
}

@inproceedings{carreira2017kinetics,
  author    = {Carreira, J. and Zisserman, A.},
  title     = {Quo vadis, action recognition? A new model and the {K}inetics dataset},
  booktitle = {Proceedings of the IEEE Conference on Computer Vision and Pattern Recognition},
  year      = {2017},
  pages     = {4724--4733}
}

@inproceedings{xu2022gdlt,
  author    = {Xu, A. and Zeng, L.-A. and Zheng, W.-S.},
  title     = {Likert scoring with grade decoupling for long-term action assessment},
  booktitle = {Proceedings of the IEEE/CVF Conference on Computer Vision and Pattern Recognition},
  year      = {2022},
  pages     = {3222--3231}
}

@inproceedings{yu2021core,
  author    = {Yu, X. and Rao, Y. and Zhao, W. and Lu, J. and Zhou, J.},
  title     = {Group-aware contrastive regression for action quality assessment},
  booktitle = {Proceedings of the IEEE/CVF International Conference on Computer Vision},
  year      = {2021},
  pages     = {7899--7908}
}

@inproceedings{xu2022finediving,
  author    = {Xu, J. and Rao, Y. and Yu, X. and Chen, G. and Zhou, J. and Lu, J.},
  title     = {{FineDiving}: A fine-grained dataset for procedure-aware action quality assessment},
  booktitle = {Proceedings of the IEEE/CVF Conference on Computer Vision and Pattern Recognition},
  year      = {2022},
  pages     = {2949--2958}
}

@inproceedings{parmar2019mtl,
  author    = {Parmar, P. and Morris, B. T.},
  title     = {What and how well you performed? {A} multitask learning approach to action quality assessment},
  booktitle = {Proceedings of the IEEE/CVF Conference on Computer Vision and Pattern Recognition},
  year      = {2019},
  pages     = {304--313}
}

\end{document}